\documentclass[english,article,oneside]{book}

\usepackage{arxiv}
\usepackage[T1]{fontenc}    
\usepackage{hyperref}       
\usepackage{url}            
\usepackage{booktabs}       
\usepackage{amsfonts}       
\usepackage{nicefrac}       
\usepackage{microtype}      
\usepackage{lipsum}
\usepackage{subfig}
\usepackage{setspace}
\usepackage{footmisc}
\usepackage{svg}
\usepackage{float}

\hypersetup{
    colorlinks = true,
    linkcolor = black,
    filecolor = black,
    urlcolor = black,
    citecolor = black,
    linkbordercolor = {0 0 1}
}

\usepackage{amsmath}  
\usepackage{booktabs}      
\usepackage{tabularx}
\usepackage{tcolorbox}
\usepackage[colorinlistoftodos, textwidth=40mm, textsize=scriptsize]{todonotes}
\usepackage{verbatim}
\usepackage{graphicx}  
\usepackage{placeins}
\usepackage{hyperref}

\usepackage[style=verbose-inote,citestyle=authoryear, uniquelist=true, uniquename=false,maxbibnames=9,maxcitenames=2,backend=biber, natbib]{biblatex}

\bibliography{references.bib}

\tikzstyle{dgraph}=[->, line width=1.5pt]
\tikzstyle{var}=[]
\tikzstyle{comp}=[]
\tikzstyle{ellipse}=[circle]

\title{Machine Psychology}

\author{
  Thilo Hagendorff\thanks{Shared first authorship. Contact: \href{mailto:thilo.hagendorff@iris.uni-stuttgart.de}{thilo.hagendorff@iris.uni-stuttgart.de}, \href{mailto:idg@google.com}{idg@google.com}} \\
  University of Stuttgart \\
  \And
  Ishita Dasgupta\footnotemark[1] \\
  Google DeepMind \\
  \And
  Marcel Binz\thanks{Co-authors are listed in alphabetical order.} \\
  Helmholtz Institute for \\Human-Centered AI \\
  \And
  Stephanie C.Y. Chan\footnotemark[2] \\
  Google DeepMind \\
  \And
  Andrew Lampinen\footnotemark[2] \\
  Google DeepMind \\
  \And
  Jane X. Wang\footnotemark[2] \\
  Google DeepMind \\
  \And
  Zeynep Akata \\
  TU Munich \\
  \And
  Eric Schulz \\
  Helmholtz Institute for \\Human-Centered AI \\
}

\begin{document}
\maketitle

\setlength{\skip\footins}{0.5cm}
\setlength{\footnotesep}{0.5cm}

\begin{abstract}
Large language models (LLMs) show increasingly advanced emergent capabilities and are being incorporated across various societal domains. Understanding their behavior and reasoning abilities therefore holds significant importance. We argue that a fruitful direction for research is engaging LLMs in behavioral experiments inspired by psychology that have traditionally been aimed at understanding human cognition and behavior. In this article, we highlight and summarize theoretical perspectives, experimental paradigms, and computational analysis techniques that this approach brings to the table. It paves the way for a "machine psychology" for generative artificial intelligence (AI) that goes beyond performance benchmarks and focuses instead on computational insights that move us toward a better understanding and discovery of emergent abilities and behavioral patterns in LLMs. We review existing work taking this approach, synthesize best practices, and highlight promising future directions. We also highlight the important caveats of applying methodologies designed for understanding humans to machines. We posit that leveraging tools from experimental psychology to study AI will become increasingly valuable as models evolve to be more powerful, opaque, multi-modal, and integrated into complex real-world settings.
\end{abstract}

\maketitle
\section*{Introduction}

Recent advances in computing power, data availability, and machine learning algorithms have yielded powerful artificial intelligence systems that are used in almost all parts of society. Among these, large language models (LLMs), gigantic neural network architectures trained on large amounts of text, have seen a particularly meteoric rise in their influence. The ability of LLMs to interface directly with natural language has made them accessible to the public in a way that was not seen before, leading to widespread adoption with millions of daily users \citep{geminiteam2024gemini,Anthropic.2024,OpenAI.2022b,OpenAI.2023b}. Also contributing to their rise in influence is that LLMs are wide-ranging in the kinds of tasks they can do -- from writing text or code to calling functions, accessing the Internet, retrieving external information, reasoning about complex problems, and many more \citep{bubeck2023sparks,lo2022gpoet,elkins2020can}. Recently, LLMs have been extended to interact with other modalities such as vision and speech \citep{fei2022towards,radford2023robust}. The ever-growing capabilities of these systems make them challenging but also increasingly important to characterize and understand, especially since these expanding capabilities also bring greater potential for unforeseen harm \citep{bommasani2021opportunities,Hagendorff.2024, weidinger2022, bender2021dangers, schramowski2022large}.

\begin{figure}[h]
    \centering
    \includegraphics[width=\textwidth]{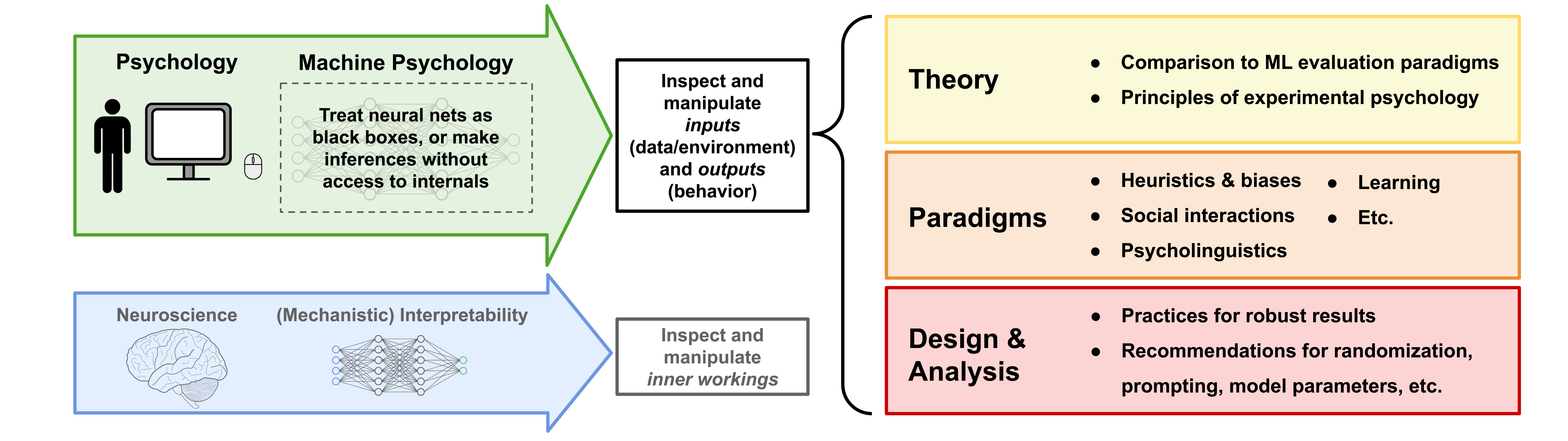}
    \vspace{0.1cm}
    \caption{Overview of key concepts of machine psychology.}
    \label{fig:figure1}
\end{figure}

Understanding behavioral patterns and emergent abilities in LLMs requires explaining their operating principles. Of the approaches focused on explaining AI systems, many rely on trying to understand the inner workings of these neural networks. This approach, often termed mechanistic interpretability, seeks to investigate LLMs by analyzing how their weights and activation patterns implement the observable behavior. It uses simplifications in terms of data, the model, or both, that make causal interventions possible and the internal mechanisms easier to characterize \citep{stolfo2023mechanistic, conmy_towards_2023, wang_interpretability_2022, gao2024scaling}. A related set of approaches draws inspiration more directly from neuroscience to characterize broader correlational similarities and differences between the internal processing of LLMs and humans \citep{hosseini2023large, kumar2022reconstructing}. 

In contrast, this review focuses on the class of approaches that directly study the \textit{behavior} of LLMs, analyzing relationships between inputs and outputs instead of inspecting the inner workings. This approach includes not only analyses of static trained models, but also experimental manipulations of inputs both during and after training. It also encompasses analyses of inputs and outputs that reveal insights about internal mechanisms, even if those internal mechanisms are not directly inspected. For this set of approaches, experiments can be inspired by human psychology, cognitive science, and the behavioral sciences. This is what we want to term \emph{machine psychology} (see \hyperref[fig:figure1]{Figure \ref{fig:figure1}}). Over several decades, the mentioned disciplines have developed a wide range of methods and frameworks to understand and characterize observable intelligent behaviors in human and non-human animals \citep{edwards1954statistical,festinger1953research}, much of which can now be adapted to LLMs as well.

Thus far, the research community has responded to the challenges of understanding behavioral patterns and growing capabilities in LLMs in several ways \citep{schwartz2022should, zhao2023explainability}. The traditional machine learning benchmark-driven approach has released new datasets that capture specific aspects only recently seen emerging in models \citep{srivastava2022, hendrycks2021measuring, zellers2019hellaswag}. Traditional benchmarking aims primarily to enable the community to compare and optimize LLM performance. In contrast, machine psychology research is not primarily interested in increasing (or measuring) an LLM’s performance, but rather in understanding behavioral patterns. While traditional natural language processing benchmarks measure abilities such as translation, numerical reasoning, or factual accuracy, machine psychology is also interested in how these observable abilities indirectly reflect the underlying constructs and algorithms \citep{experimentology2024}. Understanding these constructs lets us make new predictions about e.g. how the model will generalize, how it will perform with different training data, and specific failure modes.

The relative importance of behavior-based inspection (or psychology) versus internal inspection (or neuroscience) has been a long-standing debate \citep{jonas_could_2017}. We believe that both approaches have value for understanding both humans and LLMs. Directly inspecting LLMs' behavior, however, does come with multiple advantages. The behavior of LLMs is expressed at the interface of the model, where human users interact, and thus is what we ultimately care about the most \citep{binz2023, chang2024language, ivanova2023running}. Such behavior is often too complex to predict purely from our current mechanistic understanding of model weights and activation patterns \citep{gron2003alike}. Many interesting behaviors are only displayed by large models with billions of parameters \citep{kaplan2020scaling, wei_emergent_2022}, and behavioral methods in psychology that treat behavior directly as the experimental variable of interest scale gracefully with model size. Another practical advantage is that these behavioral approaches can easily be applied by the broader academic community to closed-source state-of-the-art models whose internal workings are not disclosed to the public.

In this article, we review and chart future directions in this emerging field of directly modeling LLM behavior. We outline how established behavioral sciences can guide and inform our understanding of LLMs, and discuss important caveats for when and how to apply methods to LLMs, given that they were originally developed for humans and animals. In the first section, we discuss the theoretical frameworks developed and used in psychology to organize our understanding of intelligence and intelligent behaviors. We then review the many empirical paradigms that have been developed to study and characterize different aspects of intelligent behavior. Finally, we discuss and make recommendations for robust empirical methods both for designing experiments and analyzing behavioral data. We end the article by discussing the potentials and limitations of conducting machine psychology experiments with increasingly capable black-box models.

\section*{Theory: Evaluation paradigms for understanding intelligent systems}
\label{sec:theory}

The traditional framework in machine learning algorithms has revolved around benchmark datasets \citep{bowman2015large, russakovsky2015imagenet}. These datasets are designed to require specific capabilities (e.g. object recognition, sentiment analysis, etc.) for good performance. Researchers train on a train dataset and evaluate on a held-out test dataset that was not seen during training. This framework does not generalize well to large-scale foundation models for two reasons. First, when using Internet-scale training data for models, this split has become harder to maintain \citep{li_task_2023, khan_xcodeeval_2023}. Second, foundation models are only directly trained for next-token prediction but exhibit many other "intelligent" behaviors that can, with some reservations \citep{schaeffer_are_2023}, be considered emergent. For example, practitioners did not explicitly encode or train for a transformer LLM's ability to learn from a few examples in context \citep{brown2020language}, but it nonetheless arose from the machine learning architecture, data, and learning signal \citep{chan2022data, von_oswald_transformers_2022}. Emergent behaviors can be difficult to study through the lens of the components that gave rise to it \citep{anderson1972more}, and the ones that emerge can seem surprising \citep{wei_emergent_2022} -- the most interesting evaluations are not `held-out' exemplars of the training task. 

Researchers have therefore started building test-only benchmarks -- i.e. smaller scale datasets unsuitable for training and intended solely as a test set -- to investigate model capabilities, e.g. the BIG-bench comprising more than 200 tests \citep{srivastava2022}, the Abstraction and Reasoning Challenge \citep{chollet2020abstraction}, as well as many others \citep{ivanova2024elements, mazumder2024dataperf}. In several cases, these benchmarks already resemble evaluation frameworks from the behavioral sciences \citep{bubeck2023sparks} -- like personality tests, intelligence tests, implicit association tests, etc. that are applied to humans -- which similarly do not follow the train-test paradigm. They also tend to fall into two categories. Some evaluations focus on scalar performance metrics, e.g. intelligence quotients. Others focus on \emph{characterizing} behavior, i.e. the questions are not designed with accuracy in mind, but designed to elicit responses that reveal behavioral strategies, or underlying constructs. In this review, we focus on test-only evaluations that provide this latter kind of understanding, as a novel evaluation paradigm that is starting to gain traction in the machine learning field.

Several such diagnostic evaluations have been developed even for pre-LLM models where, despite the models being trained for specific tasks, \textit{how} to solve them is not specified. Such diagnostic datasets were used to expose the \textit{ways} in which learned systems solved tasks -- often counter to human intuitions \citep{geirhos_shortcut_2020, mccoy-etal-2019-right, hermann_what_2020, dasgupta2022, singla_salient_2021}. Researchers have also made the case for borrowing from ethology, a branch of zoology that studies the behavior of non-human animals, to explain machine behavior in machine learning systems \citep{rahwan2019}. However, in the era of LLMs, not only are the \textit{how} unspecified, but the model abilities themselves are neither directly known nor intentionally engineered. Furthermore, since LLMs can be evaluated via natural language, this can enhance or replace comparatively simpler methods from ethology. This has led to the widespread adoption of language-based diagnostic evaluations, making it easier and more intuitive for practitioners to develop relevant tests.

However, this comes with important caveats. In trying to shed light on the workings of a black-box system that can produce language, it is tempting to use the simplest approach of asking the system about it. Self-report measures have been extensively used in psychology as well; but their reliability is questionable in humans \citep{jobe2003cognitive} as well as LLMs. Properties that such measures usually consider, such as personality, morality, or clinical disorders, are famously sensitive to prompting \citep{DominguezOlmedo.2023, röttger2024political}; to the extent that several recent works even simulate groups of humans of different social groups, opinions, and personalities with differently prompted LLMs \citep{salewski2023incontext, park2022social, argyle2023out, shanahan2023role}. There remains value in using self-report stimuli from psychology -- for example, to characterize behavior on a default prompt, as well as to understand how steerable (i.e. sensitive to prompting) models are along these dimensions. But results drawn from these measures should be taken contextually (e.g. as a property of a specific system prompt on a model) instead of as a fundamental or general property of the LLM itself. 

In contrast, the empirical tradition in psychology is significantly different from self-reports. This tradition has yielded lasting understanding of natural intelligence \citep{experimentology2024}, and is the tradition we argue is the most amenable for transferring insights to machine psychology. In this paradigm, externally observed behavior continues to be the measured experimental variable, but stimuli are designed such that different observed behaviors map onto and measure different internal representations, capabilities, or constructs -- like compositionality, theory of mind, logic, causality, etc. A key principle is that experiments are hypothesis-driven: if the agent has representation or construct X, we would expect to see behavior Y, otherwise we would see behavior Z. We highlight two key principles from this tradition that are crucial to keep in mind when performing and interpreting machine psychology evaluations. First, does seeing behavior Y reliably imply having the construct X? To answer this, the design of a good control is crucial -- to ensure that behavior Y does not have another explanation and does, in fact, implicate X. A large part of experimental psychology has been coming up with the right controls for these subtle constructs \citep{boring1954nature}, and has been providing a valuable foundation for future research in machine psychology. Second, does the absence of behavior Y indicate the absence of the construct X? This is a more subtle question. Research in psychology often grapples with the fact that human performance can be noisy or biased; for example, humans may make mistakes even on an easy calculation, or produce ungrammatical language colloquially. These should not be taken to mean that they lack the abstract capability for math or language. These inconsistencies led to the concept of the \emph{performance-competence distinction} \citep[e.g.][]{chomsky2014aspects}: that the way humans \emph{perform} in a particular situation may not fully capture their underlying \emph{competence}. More recent work has suggested that similar issues apply when assessing the capabilities of machine learning systems \citep{firestone2020performance}, and particularly LLMs \citep{lampinen2022can}. 

\section*{Paradigms: The many aspects of intelligent behavior}

There are many aspects of intelligent behavior, each of which has been studied by different sub-fields of the behavioral sciences. Each of these has developed domain-specific empirical paradigms. While some of these sub-fields (e.g. motor learning) and paradigms (e.g. pupillometry) are not directly transferable to LLMs since they rely on the existence of a physical body, several of these paradigms are purely linguistic and can be easily transferred. As LLMs expand in the kinds of stimuli they can interpret -- e.g. visual \citep{OpenAI.2023c, Zhang.2023b, geminiteam2024gemini} -- and the ways in which they can interact with the world -- e.g. embodiment and tool use \citep{mialon2023} --, the space of transferable paradigms increases. Humans also interact with several modalities, and the paradigms developed to understand us often compare and integrate these modalities \citep{schulze2023have} -- e.g. the Stroop test which spans vision and reading capabilities \citep{Scarpina.2017}.

In this article, we focus on language-based tests, since these are the most widely used in the current research landscape. Moreover, we believe that even in light of the growing trend toward multi-modal models, language will remain a primary modality due to its fundamental role in models' reasoning processes. We concentrate on four research areas that can inform distinct strands in machine psychology research: heuristics and biases, social interactions, the psychology of language, and learning. Apart from these four areas, there are, of course, multiple other domains of psychology that can also provide valuable paradigms for, for instance when investigating creativity in LLMs \citep{stevenson2022}, clinical psychology \citep{li2022}, moral behavior \citep{khandelwal2024moral}, and others.

\subsection*{Heuristics and biases}

The heuristics and biases framework is one of the most influential research paradigms in psychology \citep{gigerenzer2011heuristic, tversky1974judgment}. Heuristics are mental shortcuts that simplify reasoning or decision-making processes, and this field studies how such shortcuts can help explain both the successes and the biases in human behavior. The large existing literature on heuristics and biases in humans is a fertile ground for examining such shortcuts in the newest generation of LLMs -- whose capabilities now overlap more with the human abilities this literature studies. \citet{binz2023} were among the first to use this paradigm to better understand the decision-making processes of LLMs. They found that GPT-3 \citep{brown2020language} displays some of the same cognitive biases observed in people. Several other works have also been done in this vein \citep{jones2022capturing, palminteri2023studying, hagendorff2023human, macmillan2024ir, schulze2023have, hayes2024relative, coda2024cogbench}. Interestingly, there is evidence from several studies showing that, while the previous generation of models frequently exhibited human-like heuristics and biases, they have largely disappeared in the latest generation of LLMs \citep{chen2023emergence, hagendorff2023human}. The test stimuli were originally designed to be challenging for human study participants and possibly no longer challenge the growing reasoning abilities in LLMs. This could also be due to leakage into the training set -- we discuss this challenge in the section on design and analysis.

The literature on heuristics and biases also suggests that how a problem is phrased can influence how people solve it \citep{cheng1985pragmatic, tversky1981framing}. It is well-known that LLMs are also susceptible to similar manipulations. For example, \citet{dasgupta2022} have investigated whether LLMs are affected by the semantic content of logical reasoning problems using several existing tasks from the literature. They found that, like people, LLMs reason more accurately about familiar, believable, or grounded situations, compared to unfamiliar, unbelievable, or abstract problems. Likewise, \cite{schubert2024context} have shown that how LLMs learn in-context depends on the problem formulation. 

Finally, people do not simply apply arbitrary heuristics. Instead, they use heuristics that are adapted to the problems they encounter during their everyday interactions with the world \citep{todd2012ecological}. In the context of LLMs, one can look at how the properties of the training data shape their behavior. For example, \cite{chan2022data} have demonstrated that the presence of in-context learning in LLMs can be traced back to data distributional properties such as burstiness, where items appear in clusters rather than being uniformly distributed over time, and the presence of large numbers of rarely occurring classes. Researchers also proposed that one should try to understand LLMs through the problem they are trained to solve, similarly to how behavioral scientists attempt to understand human cognition through the lens of ecological rationality \citep{todd2012ecological, mccoy2023embers, jagadish2024ecologically}. 

\subsection*{Social interactions}

Traditionally, developmental psychology explores how humans develop cognitively, socially, and emotionally throughout their lives. This includes studying the various factors that influence development, such as social intelligence or social skills. By applying paradigms from this area of developmental psychology to LLMs, researchers can gain deeper insights into how these models manage complex social interactions. In particular, once LLMs are deployed as chat agents, they should become versed in modeling human communicators. Therefore, it is important to assess the level of social intelligence in LLMs. One example in this context is the application of theory of mind tests to LLMs, where researchers use tasks from human experiments, such as those famously conducted by \citet{wimmer1983} and \citet{Perner.1987}. While early experiments with models such as GPT-3 showed that they struggle to solve theory of mind tasks \citep{Sap.2022}, later models demonstrate an increasing ability to reliably infer unobservable mental states in others \citep{Strachan.2024, holterman2023, Moghaddam.2023}. Further related research examines how LLM performance on theory of mind tests compares to that of children \citep{vanDuijn.2023}, LLM ability to handle higher-order theory of mind tasks requiring recursive reasoning about multiple mental states \citep{Street.2024}, or measures the robustness of theory of mind test setups against distracting alterations in the tasks LLMs receive as inputs \citep{ullman2023}. As theory of mind tests measure, among other things, the ability to understand false beliefs, further research has explored the emerging capability of LLMs to induce false beliefs in other agents \citep{Hagendorff.2024b}, or how LLMs trade off various communicative values like honesty and helpfulness \citep{liu2024large} -- these investigations also contribute to understanding and improving alignment with human values for AI safety \citep{ji2024ai}.

The space of relevant paradigms increases as LLMs are allowed to interact through self-reflection \citep{Nair.2023}, self-instruction \citep{Wang.2022b}, or in swarms \citep{Zhuge.2023}. For example, researchers looked at cooperative and coordinative behavior in LLMs playing games, revealing persistent behavioral signatures in the models \citep{Akata.2023}. Similarly, researchers investigated cooperative or competitive LLMs behavior in psychology-inspired dilemma situations to assess the ability of LLMs to participate in real-world negotiations \citep{phelps2024machinepsychologycooperationgpt}. Another study, which is influenced by works in human social psychology, looked at how multiple LLMs form and evolve networks, investigating micro-level network principles such as preferential attachment or triadic closure, as well as macro-level principles such as community structures \citep{Papachristou.2024}. In sum, machine psychology can reveal patterns of social behavior and interaction among LLMs, individually and collectively, be it for problem solving or world simulation \citep{Guo.2024}. By drawing from human developmental psychology and social dynamics, researchers can better understand and design LLMs that navigate complex social interactions and exhibit advanced social skills.

\subsection*{Psychology of language}

A long history of work has studied the psychology of how humans use and understand language, ranging from how they use semantic and syntactic features to understand a sentence to how they use pragmatic inferences in a discourse context to help interpret what someone has said. Correspondingly, a long-standing body of work has studied how language processing models capture these features of human language processing. Early connectionist works studied these topics in simple recurrent predictive models \citep{elman1991distributed,mcclelland1989sentence}; more recently, researchers have applied similar techniques to study LLMs. A wide range of work has studied what models learn about syntax \citep{linzen2021syntactic}, often using methods from psycholinguistics. For example, \citet{wilcox2023using} used psycholinguistics-inspired surprisal measures to show that LLMs learn filler-gap dependencies, a challenging syntactic structure. Other researchers have used related measures to study what LLMs learn about the semantics of entailment \citep{merrill2024can}. Moreover, researchers used psycholinguistic techniques like priming to study how models represent and process language \citep{prasad2019using,sinclair2022structural}, and methods like deconfounded stimuli to identify where models may rely on semantic heuristics rather than syntax \citep{mccoy-etal-2019-right}. Several recent works \citep{hu2023fine,ruis2023goldilocks} studied pragmatic judgments of LLMs, and found that larger models, as well as those with instruction tuning, tend to better approximate human responses and error patterns -- though some deficiencies remain. In another study, researchers examined long-form analogies generated by ChatGPT, finding that AI-generated analogies lack some human-like psycholinguistic properties \citep{seals2023longform}, particularly in text cohesion, language, and readability. Furthermore, researchers applied garden path sentences -- sentences that lead the reader to initially interpret them incorrectly due to their ambiguous structure -- to LLMs, showing that the models respond similarly to humans \citep{aher2023using, CHRISTIANSON2001368}. At a higher level, some researchers have drawn inspiration from aspects of human language development to attempt to identify the causes of the relative data inefficiency of language models \citep{warstadt2023findings,frank2023bridging}. In each of these cases, methods and ideas from psychology and psycholinguistics provide guidance on how to assess processes through language behaviors in LLMs, potentially by drawing comparisons between LLMs and humans.

\subsection*{Learning}
The psychology of learning is concerned with how individuals acquire and retain knowledge and skills. At first blush, it may appear that experimental paradigms for the study of learning are less applicable to LLMs, given that the aim of behavioral experiments is often to help uncover the underlying learning algorithm – whereas for LLMs the learning algorithms used in training are designed and already known. However, the behavioral sciences can still benefit from the study of LLMs in this context, since LLMs exhibit learning abilities that were not explicitly designed into the models (they are emergent), and thus one does not understand the underlying learning algorithm. In particular, LLMs exhibit emergent in-context learning -- the ability to learn from context (the prompt) without requiring any gradient-based updates in weights \citep{brown2020language}. Understanding in-context learning is a burgeoning field that is rapidly gaining in importance, given the increasing size of LLMs context windows and consequent gains in capabilities, e.g. the capability to learn an entire language from context alone \citep{munkhdalai2024leave, geminiteam2024gemini}, or the ability to overcome safety fine-tuning \citep{anil2024many, zheng2024improved}. 

Uncovering the implicit learning algorithm implemented by in-context learning is a burgeoning research field, and utilizes many of the methods common in cognitive science. For example, multiple studies have compared the outputs of transformer in-context learning with the outputs of hypothesized learning algorithms \citep{von_oswald_transformers_2022, akyurek_what_2022}. This is a staple of cognitive modeling, and could potentially benefit even further from model comparison procedures from psychology and statistics \citep{yang_comparing_2006, arlot_survey_2010, vrieze_model_2012}. Recent work in cognitive science has used machine learning to discover new theories of human decision-making \citep{peterson_using_2021} -- it might be interesting to apply related approaches to in-context learning as well. Researchers might also benefit from considering particular models as normative starting points \citep{niv_reinforcement_2009}.

Researchers may also wish to understand other interesting and important characteristics of learning, such as inductive biases and generalization, the data dependence of learning, and the dynamics of learning over time. These characteristics are often not obvious even in cases where the learning algorithm is known, and thus researchers would like to understand them not only for in-context learning, but also for other forms of LLM learning, e.g. self-supervised gradient-based learning, reinforcement learning \citep{ouyang_training_2022}, or "fast" memory retrieval \citep{borgeaud_improving_2021, lewis_retrieval-augmented_2021}.

To characterize inductive biases and generalization of LLMs, researchers have borrowed both concepts and experimental paradigms from cognitive sciences \citep{schubert2024context, coda2023meta} and Bayesian inference \citep{xie2022explanation}. Studies utilized paradigms for measuring systematic generalization to characterize those capabilities in LLMs, and as inspiration to improve these abilities \citep{Lake2023,Ruis2022}. \citet{webb2022} created novel variants of classic analogy problems from cognitive science, in order to examine the analogical capabilities of large language models. \citet{chan2022data} have borrowed ideas and experimental paradigms on "rule-based" vs. "exemplar-based" generalization to characterize the inductive biases of in-weights vs. in-context learning in transformers. Furthermore, researchers borrowed paradigms and measures from developmental psychology to characterize the domains where LLM inductive biases may match those of children, and where they may fall short (including in causal reasoning and innovation) \citep{kosoy2023comparing, yiu2023transmission}.

To characterize the data dependence of in-context learning, existing work has drawn inspiration from research in developmental psychology on skewed and bursty distributions \citep{chan2022data}. An important aspect of data dependence is the structure of data over time (during training). AI researchers have long drawn inspiration from curriculum learning in human and non-human animals to better understand how to structure training data so that earlier learning on easier tasks can scaffold later learning on harder tasks \citep{10.1145/1553374.1553380}. There remain many areas of behavioral research on learning that may serve as rich sources of inspiration on data dependence, e.g. research on repetition and spacing \citep{dempster_spacing_1989}, working memory \citep{baddeley_working_1992, chai_working_2018}, blocking vs. interleaving tasks \citep{carvalho_benefits_2015}, and continual learning \citep{greco_psycholinguistics_2019}. Data dependence is particularly interesting for LLMs because text training data (being sourced largely from unstructured web-scale corpora) is very different from the structured training data typically used for traditional discriminatory machine learning techniques, and because data is one of the major levers one can manipulate in training LLMs to adjust their behaviors.

\section*{Design and analysis: Good behavioral experimentation}
Computer science has not historically been an empirical science. While machine learning (especially since the era of neural network models) has been significantly driven by empirical rather than theoretical work, the settings under which those protocols were developed -- a test set that is fixed for all practitioners and is effectively infinitely large -- no longer hold in the small test-only behavioral experiments setting. Current LLMs are famously sensitive to small changes in prompt structure or they rely on shallow syntactic heuristics \citep{mccoy-etal-2019-right}, and studies that are not careful about testing the robustness of their conclusions risk being spurious and non-generalizable. Psychology too has had its own share of reproducibility crises \citep{OpenScienceCollaboration.2015, HaibeKains.2020}, and machine psychology should not share the same fate. In this section, we provide recommendations for sound methodologies in behavioral test settings with LLMs, which should be valuable to practitioners in the field of machine psychology.

\subsection*{Prompting methods and biases}
Many studies conducted in the field of machine psychology have a significant shortcoming in common, namely that they do not avoid training data contamination. They use prompts from existing psychology studies and apply them to LLMs without changing their wording, task orders, etc. In this way, LLMs are likely to have already experienced identical or similar tasks during training, thus causing LLMs to simply reproduce known token patterns. When adopting test frameworks from psychology -- meaning vignettes, cognitive tasks, or other test setups -- researchers must ensure that LLMs have never seen the tests before and go beyond mere memorization. Hence, prompts may indeed be structurally like already existing tasks, but they should contain new wordings, agents, orders, actions, etc. That being said, some experiments may be procedurally generated (instead of consisting of a static dataset), which makes them inherently less susceptible to data contamination issues \citep{coda2024cogbench}. 

Another common shortcoming of several existing machine psychology studies is that they rely on small sample sizes or convenience samples, meaning non-systematic sequences of prompts. Sampling biases in the used benchmarks or task datasets, which are especially prevalent in small sample sizes, can diminish the quality of machine psychology studies. This is because slight changes in prompts can change model outputs significantly. Because of this high sensitivity to prompt wording, it is important to test multiple versions of one task and to create representative samples, meaning batteries of varied prompts. Only in this way can one reliably measure whether a certain behavior is systematically reoccurring and generalizable \citep{Yarkoni_2022}. Furthermore, LLMs can succumb to various biases influencing the processing of prompts \citep{zhao2021,chan2022data}. Recency biases in LLMs, for instance, lead to a tendency to rely more heavily on information appearing toward the end of prompts. LLMs can also possess a common token bias, meaning that models are biased toward outputting tokens that are common in their training data. Moreover, majority label biases can cause LLMs to be skewed towards labels, classes, or examples that are frequent in a few-shot learning setting. Technical biases like these can at least in part be controlled for when designing prompts or prompt variations that tend to avoid triggering them. If this is not done, LLMs may rely on shortcuts exploiting such biases.

\subsection*{Eliciting capabilities with prompts}
The standard prompt design, comprising a vignette plus an open- or close-ended question or task, can be enhanced by prefixes or suffixes eliciting improved reasoning capabilities in LLMs. On the other hand, omitting such prefixes and suffixes can lead to underestimations of the model's capabilities. Although it is likely that most specific prompt augmentations have a positive influence on one kind of task but not another, reducing our ability to systematically understand LLM behavior, a few prompt design approaches have nonetheless been found to confer broader performance benefits. Most notably, (zero-shot) chain-of-thought prompting \citep{Wei.2022, kojima2022} -- which simply adds “Let’s think step by step” at the end of a prompt -- improves reasoning performance. This can be extended even further by generating multiple chain-of-thought reasoning paths and taking the majority response as the final one \citep{Wang.2022b}. Similar to chain-of-thought prompting is least-to-most prompting, which also decomposes problems into a set of subproblems to increase accuracy in LLMs \citep{zhou2022}. Yet another approach is to frame questions in a multiple-choice format. This was shown to improve reasoning capabilities in some cases \citep{kadavath2022}, but can also limit them because LLMs might be prompted to provide brief responses, thereby circumventing reasoning in the process of prompt completion. Nevertheless, many prominent NLP benchmarks use multiple choice formats instead of open-ended questions. Here, one must keep in mind that different expressions of the same concept compete for probability, which can lower the chances of selecting the correct answer \citep{holtzman2022surfaceformcompetitionhighest}. Moreover, one has to consider potential recency biases, which require neutralizing this effect by shuffling the order of answers in multiple test runs to cover all possible combinations. Another method to increase reasoning is to utilize the ability for few-shot learning in LLMs \citep{brown2020language}, where the LLM's performance improves after repeated exposure to a given task. Moreover, self-reflection, meaning the automated, recursive criticizing and subsequent self-improvement of LLM outputs by the LLM itself, is a further technique that can improve reasoning abilities \citep{Nair.2023, kim2023}. Regarding improvements in symbolic or numeric reasoning, another technique is to prompt LLMs to use code for solving tasks \citep{zhang2024natural}. Eventually, all mentioned methods to improve reasoning can be not just leveraged for machine psychology; they can also become objects of study themselves.

\subsection*{Setting parameters and evaluating outputs}
LLMs come with a variety of parameters researchers can set. For example, most models come in a variety of sizes. Analyses across different sizes are valuable: while the largest ones usually have the highest capabilities, some recent works find "inverse-scaling" \citep{mckenzie2023inverse}. Moreover, temperature settings control randomness. If exact reproducibility is required, studies should use temperature 0 or assign a seed to ensure complete determinacy. However, this can be prone to (intentional or unintentional) biases in seed choice. The effect of temperature on capabilities is not established \citep{renze2024effect}, and reporting averages or "best of K" -- considering all the responses over K samples that meet certain simple criteria, e.g. formatting \citep{chen2021evaluating} -- is valuable.

After conducting the experiments, a list of LLM responses must be evaluated and compared with the ground truth. The simplest case is when the results can be framed and scored as a multiple-choice question -- though even in this case, scoring the answers so that the model responds directly inline, rather than selecting a choice, can yield more signal \citep{hu2023prompting}. If possible, multiple scoring methods should be compared, to evaluate whether the effects are dependent on the scoring method \citep{tsvilodub2024predictions}. If the questions must be answered with free generations, the evaluation process can still be automated if the results exhibit sufficient simplicity and regularity, meaning that the LLM responses are similar to the ground truth strings in terms of length and wording, which is particularly common when using masked language models. Methods such as testing word overlaps with regular expressions or using metrics such as the F1 score can be employed. State-of-the-art LLMs, however, tend to produce highly variable and comprehensive outputs, which can complicate classification. While stop sequences, token limits, or prompt instructions that interrupt further text generation can facilitate classification by promoting output uniformity, they also improperly constrain LLM behavior. Therefore, researchers are increasingly relying on LLM-based evaluations of outputs where a single model or multiple stacked model instances perform the classification using carefully crafted instructions. Although this method might still be inaccurate for very comprehensive outputs, a solution is to instruct the LLM under scrutiny to output its final answer or summary after a specific string sequence like "\#\#\#\#" \citep{cobbe2021training}. This approach allows the LLM to reason during verbose prompt completions, which is necessary for many prompt engineering techniques such as chain-of-thought reasoning. The classification then only involves processing the string following "\#\#\#\#". If this method still proves to be unreliable, evaluations might have to be performed manually, possibly by hiring research assistants or contractors. Following the evaluation, a statistical analysis can be carried out.

\section*{Discussion}
Machine psychology provides a new approach to explaining AI. Instead of interpreting a neural network’s design components \citep{BarredoArrieta.2020}, one analyzes the relationships between inputs and outputs, i.e. prompt design and prompt completion. Although this may allow the identification of hitherto unknown abilities or behavioral traits in LLMs, interpreting LLM responses comes with a challenge. A strong tendency exists to confer mental concepts or psychological terms to LLMs that were hitherto reserved for human and animal minds. This tendency manifests in common terms like "machine learning," but will become more prevalent in machine psychology when concepts such as reasoning \citep{huang2022}, intuition \citep{hagendorff2023human}, creativity \citep{stevenson2022}, intelligence \citep{webb2022}, personality \citep{miotto2022}, mental illnesses \citep{li2022}, etc. are transferred to LLMs. In this context, researchers have demanded caution by stressing that the underlying neural mechanisms for these concepts are different in humans and machines \citep{shanahan2022, mahowald2024dissociating}. Moreover, many psychological concepts are normatively laden and can foster mismatches in expectations between AI experts and the public regarding machine capabilities \citep{shevlin2019}. Nevertheless, the problem that many abilities in LLMs cannot be reasonably grasped by only referring to the inner workings of their neural architecture remains.

By adopting a concept from ethnography, one could call such an approach "thin descriptions" \citep{Ryle.2009, Geertz.2017}, meaning that one only explains internal representations in AI systems, for instance via activation atlases, which visualize how different parts of a neural network respond to various inputs \citep{Carter.2019}. In this sense, LLMs simply hijack humans’ intuitions to explain machine behavior patterns by using psychological or other anthropocentric terms. Contrary to thin descriptions, though, there are "thick descriptions." They imply using psychological terms to add a layer of explainability. LLMs are, like the human brain, black boxes to some extent. By applying psychological terms to them, the explanatory power increases, even if no direct neural correlates to these terms exist. This holds for humans, too, where mental terms used to explain behavior do not directly correlate with specific sets of neural activations. By postulating (mental) unobservable states, be it with regard to brains or artificial neural networks, one increases explanatory resources \citep{Sellars.1997}. Thick descriptions help in making sense of LLMs when thin descriptions are insufficient to explain behavioral patterns. Thin descriptions assume that LLMs merely possess syntax or a statistical capacity to associate words \citep{searle1980, Floridi.2020, bender2021dangers}, but not semantics. Thick descriptions, though, assume that LLMs show patterns and regularities that go beyond mere syntax. These patterns can be explained by means of machine psychology.

Beyond potential habituations regarding the use of terminology borrowed from psychology in the context of machines, machine psychology, as a nascent field of research, aims to identify behavioral patterns, emergent abilities, and mechanisms of decision-making and reasoning in LLMs by treating them as participants in psychology experiments. This new discipline of evaluating LLMs will become even more important when taking multimodal or augmented LLMs into account, meaning LLMs that are allowed to interact with images, external information sources, sensory data, physical objects, and various other tools \citep{mialon2023, Schick.2023, ma2024dreureka}. Moreover, once test settings for machine psychology are established, researchers can investigate how LLMs develop over time by applying the same tasks multiple times, yielding longitudinal data. This data can serve as a baseline to extrapolate trends regarding the development of reasoning abilities in LLMs. Such estimations may be increasingly important for AI safety and AI alignment research to predict future behavioral potentials in LLMs. By gaining a deeper understanding of these potentials, machine psychology is providing a new approach to AI explainability as well as an important addition to traditional benchmarking methods in natural language processing.

\section*{Author contributions}
TH and ID conceptualized and led the initial design of the manuscript. TH and ID wrote the initial drafts, with contributions from MB, SCYC, AL, JW, ZA, and ES to flesh out the sections and create the figure. All authors assisted with iterations and edited and reviewed the paper.

\printbibliography[title=References, heading=subbibliography]

\end{document}